\documentclass[runningheads]{llncs}
\usepackage[T1]{fontenc}
\usepackage{graphicx}
\usepackage{booktabs}  
\usepackage{graphicx}
\usepackage{caption}
\usepackage{subcaption}  
\usepackage{hyperref}       
\usepackage{url}            
\usepackage{booktabs}       
\usepackage{amsfonts}       
\usepackage{nicefrac} 
\usepackage{pifont} 
\usepackage{microtype}      
\usepackage{subcaption}
\usepackage{todonotes}
\usepackage{amssymb}
\usepackage{amsmath}
\usepackage{multirow}
\usepackage{array}
\usepackage{xcolor}

\definecolor{futuretask}{RGB}{150,150,150}

\newcommand{\dataSynA}[0]{\ensuremath{\mathcal{D}_\text{1}}}
\newcommand{\dataSynB}[0]{\ensuremath{\mathcal{D}_\text{2}}}
\newcommand{\dataSynC}[0]{\ensuremath{\mathcal{D}_\text{3}}}
\newcommand{\dataSynD}[0]{\ensuremath{\mathcal{D}_\text{4}}}

\newcommand{\dataEnedisResidentiel}[0]{\ensuremath{\mathcal{D}_\text{res}}}
\newcommand{\dataEnedisEolien}[0]{\ensuremath{\mathcal{D}_\text{wind}}}
\newcommand{\dataEnedisPhoto}[0]{\ensuremath{\mathcal{D}_\text{photo}}}

\newcommand{\dataElisa}[0]{\ensuremath{\mathcal{D}_\text{ind}}}

\newcommand{\scenarioSynthetic}[0]{\ensuremath{\mathcal{S}_\text{syn}}}
\newcommand{\scenarioReal}[0]{\ensuremath{\mathcal{S}_\text{energy}}}
\newcommand{\tabSectionName}[1]{\textbf{#1}}

\begin{document}
%


\title{Foundation vs. Specialized Models: Evaluating Catastrophic Forgetting in Continual Time Series Forecasting}
\titlerunning{TSFM vs. small models in continual TSF}
%

\author{Nouha Karaouli\inst{1} \and Denis Coquenet\inst{1}\orcidID{0000-0001-5203-9423} \and
Elisa Fromont\inst{1}\orcidID{0000-0003-0133-3491} \and Martial Mermillod\inst{2}\orcidID{0000-0003-4367-7049} \and Marina Reyboz\inst{3}\orcidID{0000-0002-3373-2908}}
\authorrunning{Karaouli et al.}
%
\institute{Univ. Rennes, CNRS, Inria, IRISA - UMR 6074, F-35000 Rennes, France \and
Univ. Grenoble Alpes, Univ. Savoie Mont Blanc, CNRS, LPNC, Grenoble, France \and Univ. Grenoble Alpes, CEA, LIST, 38000 Grenoble, France}
\maketitle              

\begin{abstract}
While Time Series Foundation Models (TSFMs) excel in zero-shot tasks, their behavior under continual fine tuning is poorly understood. We present the first systematic study of catastrophic forgetting in TSFMs (TimesFM-2.0, Chronos-2) versus a specialized SamFormer model across synthetic and real-world energy forecasting benchmarks. Our results show that while fine-tuning improves new task accuracy, it consistently triggers forgetting, though larger models exhibit greater inherent robustness. Notably, employing forgetting mitigation techniques such as DER, levels the playing field: it provides disproportionate gains to smaller models, allowing them to match TSFM performance by the end of the continual learning sequence. These findings suggest that in realistic, non-stationary scenarios, the high computational cost of large foundation models may not be justified over smaller models equipped with effective mitigation strategies.

\end{abstract}

\section{Introduction}

Foundation models have revolutionized Natural Language Processing (NLP) and Computer Vision (CV), achieving strong zero-shot and few-shot generalization through large-scale self-supervised pretraining.
In NLP, models such as GPT \cite{radford2018improving} and BERT \cite{devlin2019bert} have demonstrated remarkable transfer capabilities \cite{devlin2019bert}\footnote{\url{https://github.com/openai/gpt-2}}.
In CV, large vision-language and vision-only models have similarly advanced zero-shot learning \cite{radford2021learning,yu2022coca,schneider2024foundation}. Despite their success, these models are prone to catastrophic forgetting \cite{Shi26,kirkpatrick2017overcoming}, \textit{i.e.}, the tendency to lose previously acquired knowledge when sequentially fine-tuned on new tasks \cite{ramasesh2022effect,haque2025catastrophic,zhai2023investigating}. Studies in NLP and CV showed that model scale, architecture, and pretraining significantly influence the degree of forgetting, with larger pretrained models often exhibiting more robust knowledge retention \cite{luo2025empirical,ramasesh2022effect}.

Inspired by these developments, Time Series Foundation Models (TSFMs) have emerged as a promising way to bring the benefits of large-scale pretraining to time series forecasting, enabling zero-shot and few-shot forecasting.

While these models can achieve reasonable zero-shot performance on simple datasets, they often struggle to produce accurate forecasts on specific, unseen datasets \cite{karaouli2025foundational}. This stems from the intrinsic challenges of time series data, namely continuity, noise, and non-stationarity, which make generalization difficult \cite{gruver2023large,zeng2022transformers,zhou2021informer}. As a result, fine-tuning (FT) is necessary to adapt pretrained models to new domains. However, FT may introduce the risk of catastrophic forgetting in TSFMs, whose adaption to a new dataset can degrade performance on previously learned ones.


To investigate this issue, we (i) propose a task-incremental continual learning (CL) scenario for energy forecasting that is deliberately designed to induce catastrophic forgetting across successive tasks, (ii) introduce a model-aware transferability measure to anticipate forgetting by quantifying cross-dataset knowledge transfer, and (iii) conduct a systematic assessment of catastrophic forgetting in TSFMs within a CL framework. We evaluate two recent TSFMs, TimesFM \cite{das2024timesfm} and two variants of Chronos \cite{ansari2025chronos2} (Chronos‑2 and Chronos‑small), spanning different architectures and scales, and we compare them against a state‑of‑the‑art task‑specific model, Samformer \cite{ilbert2024samformer}, on two CL scenarios. Finally, we study how these models perform in a continual learning setting when using Dark Experience Replay (DER) \cite{der2020}, a widely adopted replay-based strategy for mitigating catastrophic forgetting, recently explored for large language models (LLMs) in \cite{Shi26}.
The remainder of this paper is organized as follows. Section~\ref{sec:related} reviews the related work. Section~\ref{sec:scenario} presents the continual learning framework and evaluation methodology. Section~\ref{sec:expe} reports and analyzes the experimental results. Finally, Section~\ref{sec:conclu} concludes the paper.
  
\section{Related Works}
\label{sec:related}

\subsection{Continual Learning}
Continual Learning (CL) is a paradigm designed to enable machine learning models to learn sequentially from a stream of tasks while mitigating catastrophic forgetting \cite{zhai2023investigating}, i.e., the tendency of a model to lose performance on previously learned tasks when trained on new ones. In the literature \cite{wang2024comprehensive}, CL methods are generally grouped into three main categories: (i) regularization-based approaches which constrain parameter updates to preserve knowledge from past tasks; (ii) parameter expansion methods which increase model capacity to accommodate new tasks; and (iii) replay-based methods, such as DER \cite{der2020}, which implicitly or explicitly store a subset of past data in a memory buffer and replay it during training to maintain performance on previous tasks.

In the following, we will focus on DER as it is an intuitive and fast-implemented method that can be easily adapted to all base models as also done in \cite{Shi26} for large language models (LLMs). DER mitigates catastrophic forgetting by combining experience replay with prediction distillation. This method maintains a fixed-size buffer of past inputs and the model’s predicted outputs (“logits”), and employs reservoir sampling \cite{reservoir} to fill the buffer, ensuring that a representative subset of past data is stored. During training on a new task, a replay loss encourages the model to reproduce the stored predictions while simultaneously optimizing for the current task’s true future values. In our work, we apply DER to time series forecasting, where the buffer stores past context sequences and their predicted future horizons. 

\subsection{Continual Learning on Time Series}
In the context of time series, continual learning has recently gained increasing attention due to the inherent non-stationarity and evolving distributions of real-world temporal data. Several studies have adapted CL techniques to forecasting tasks and highlighted important limitations of classical approaches. For instance, the authors of \cite{ao2023continual} show that regularization-based methods like EWC and LwF are often insufficient, as they primarily improve stability but fail to effectively accumulate knowledge across tasks. Similarly, \cite{matteoni2022continual} demonstrate that replay-based techniques are significantly more effective for mitigating forgetting in time series settings. Recent empirical analyses \cite{Bayle2024}, do not consider any foundation models, but further confirm that methods such as DER provide strong performance in continual time series forecasting scenarios. 

To our knowledge, the closest study to ours is \cite{Liu25}; however, it differs on two main points. 
First, their analysis is limited to continual fine-tuning and does not explore state-of-the-art approaches for mitigating catastrophic forgetting. 
Second, they reach the opposite conclusion regarding the usefulness of TSFM for forecasting in a CL setting. We assume that their CL scenario/datasets may not exhibit a substantial distribution shift capable of degrading TSFM performance in CL.
This survey \cite{besnard2024continual}, though recent, does not consider state-of-the-art methods out of regularization methods which go against what is proposed in \cite{Gunasekara2023}.
In our work, we carefully design both synthetic and real CL scenarios to induce substantial temporal data shifts. For the real-world CL scenario, we ensure that at least one forecasting task is based on private data that could not have contaminated the TSFM pre-training corpus since contamination would artificially increase the TSFM performance. 

\subsection{Time Series Foundation Models}
\label{sec:TSFM}
Inspired by the success of foundation models for Natural Language Processing (NLP) and Computer Vision (CV), TSFMs have emerged to generalize across forecasting tasks, domains, and time scales. Notable examples include TimesFM \cite{das2024timesfm}, Chronos and Chronos 2 \cite{ansari2024chronos}, Tirex \cite{auerTiRexZeroShotForecasting2025}, Moirai \cite{woo2024unified} and Granite-Flowstate \cite{graf2025flowstate}. These models leverage transformer-based or attention-driven architectures to capture long-range temporal dependencies, achieving strong zero-shot and FT performance across diverse benchmarks \cite{Liang24}. Benchmarking frameworks such as GIFT-eval \cite{aksu2024gift}, OpenTS \footnote{\url{https://decisionintelligence.github.io/OpenTS/}}, and Nixtla Arena \footnote{\url{https://github.com/Nixtla/nixtla/tree/main/experiments/foundation-time-series-arena}} provide standardized evaluations of TSFMs, but they focus on offline settings \cite{van2019three} and do not focus on any continual learning framework, a critical aspect for real-world continual forecasting.

In the following, we study the performance of two recent TSFMs: TimesFM \cite{das2024timesfm}, and 2 versions of Chronos-2 \cite{ansari2024chronos}: Chronos-2 and chronos-small. This selection was performed based on their rank on previously mentioned benchmarks and on the availability of source code and trained weights. \textbf{Chronos-2} and \textbf{Chronos-small} are encoder-only transformers with group and time attention layers, pretrained on univariate and synthetic multivariate datasets including TSI \cite{bahrpeyma2021methodology} and TCM \cite{runge2023causal} generators. Chronos-2 and Chronos-small include $120M$ and $28M$ parameters respectively. \textbf{TimesFM-2.0} is a decoder-only transformer designed for long-horizon forecasting, pretrained on a mixture of real (Google Trends, Wiki Pageviews, M4, Traffic, Electricity, Weather) and synthetic time series, including $500M$ parameters. 

\subsection{Time Series Forecasting}
Contrarily to the trend of increasing model scale, recent work has demonstrated that "shallow" models can achieve competitive, and often superior, results on time series forecasting. A notable example is SAMformer \cite{ilbert2024samformer}, which utilizes a lightweight Transformer structure with about $100K$ parameters, enhanced by Sharpness-Aware Minimization (SAM) and channel-wise attention. By prioritizing a flatter loss landscape and efficient cross-variable modeling, SAMformer consistently matches or exceeds the zero-shot performance of the much larger foundation models presented in Section \ref{sec:TSFM}. In the following, we use Samformer as our baseline for continual learning against TSFMs.

\section{Experimental framework}
\label{sec:scenario}

\subsection{Continual Learning Scenario}

We define a continual learning scenario $\mathcal{S}$ as a sequence of $N$ datasets ($\mathcal{D}_1, \ldots, \mathcal{D}_N$) that are successively considered for training. In this work, each dataset $\mathcal{D}_i$ is a univariate time series implying $T_i$ samples, and corresponds to a distinct stage in the continual learning framework:
\[
D_i = \{(x_k^{(i)}, y_k^{(i)})\}_{k=1}^{T_i},
\] 
with $x_i^{(t)} \in \mathbb{R}^L$ is a context window and $y_i^{(t)} \in \mathbb{R}^H$ is the forecast horizon.
After each stage, the model performance is evaluated on all previously encountered datasets to quantify catastrophic forgetting, as well as on the current dataset to assess adaptability.

We instantiate this framework in two independent experimental settings, each containing 4 datasets: 
(i) a controlled, synthetic scenario $\scenarioSynthetic = (\dataSynA,$ $\dataSynB, \dataSynC, \dataSynD)$, and 
(ii) a real-world scenario focusing on energy consumption predictions $\scenarioReal = (\dataEnedisResidentiel, \dataEnedisPhoto, \dataElisa, \dataEnedisEolien)$. 
The two scenarios are strictly independent, with no interaction between synthetic and real-world datasets.

\subsubsection{Synthetic scenario \scenarioSynthetic}

\begin{figure}[t]
    \centering
    \includegraphics[width=0.9\linewidth]{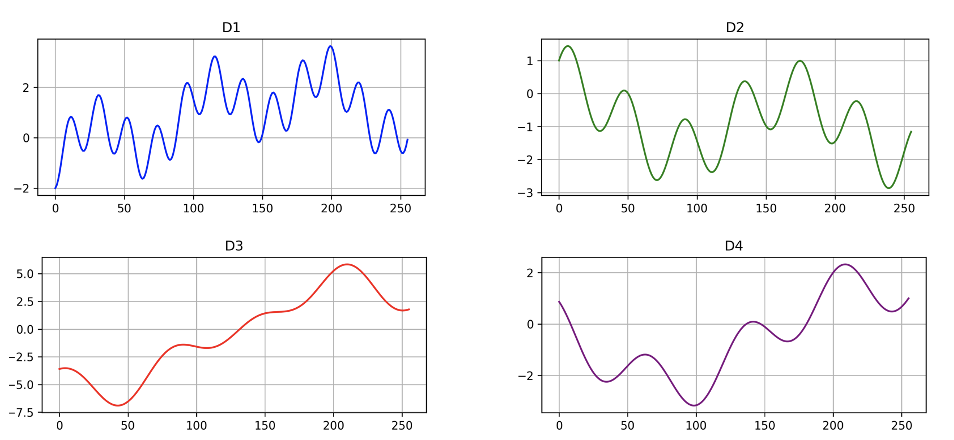}
    \caption{Typical examples of time series in the 4 synthetic datasets (D1 to D4).
    \label{fig:synthData}}
\end{figure}

The {\emph synthetic} scenario consists of four multi-sinusoidal time series datasets of increasing complexity and periodic structure, with each dataset containing 2,688 time steps:
\begin{itemize}
    \item \textbf{$\dataSynA$} and \textbf{$\dataSynB$} feature 4 and 3 sine waves respectively, with harmonically aligned periods, allowing the model to observe full cycles within the dataset. This setup tests the ability to learn and predict fully repetitive patterns.

    \item \textbf{$\dataSynC$} and \textbf{$\dataSynD$} contain 10 sine waves with randomly sampled, non-harmonic periods, producing very long global cycles that exceed the dataset length. These datasets simulate real-world scenarios where only partial signal cycles are observed, challenging the model to generalize and extrapolate from incomplete information. These scenarios allow us to test the model's ability to discover the underlying function that can predict the future steps of the time series.
\end{itemize}
Figure \ref{fig:synthData} shows typical examples of time series in the 4 synthetic datasets D1 to D4. 

\subsubsection{Real scenario \scenarioReal}

\begin{figure}[t]
    \centering
    \includegraphics[width=0.9\linewidth]{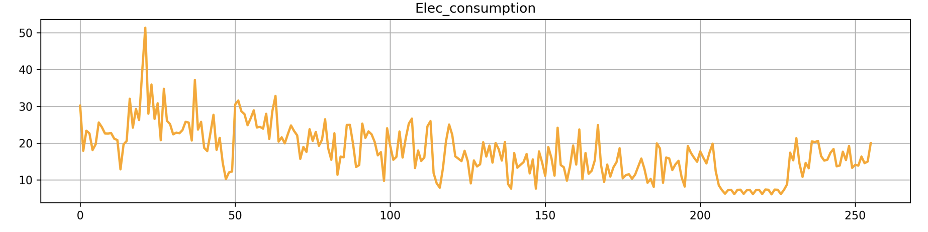}
    \caption{A typical electrical consumption series from the ENEDIS open data repository.}
    \label{fig:realData}
\end{figure}

The real‑world CL setting comprises four datasets, all sourced from Enedis, the principal national electricity provider in France, which is committed to open‑data sharing \footnote{\url{https://opendata.enedis.fr/datasets/}}. Since these data are publicly accessible, they could have been employed to pre‑train a TSFM. To guarantee that at least one scenario remains unseen by all recent TSFM models, we incorporated (as the third dataset in the sequence) the individual electricity consumption of a particular household that Enedis does not disclose publicly. The time‑series contained in each of the four datasets are all rooted in the same energy‑systems domain, yet they exhibit markedly distinct temporal dynamics. Some series correspond to aggregated consumption at the sub‑station level (one example is given in Fig. \ref{fig:realData}), others to individual household demand, while the remaining two describe renewable‑generation profiles (solar and wind). Consequently, the data differ in sampling frequency, seasonality, volatility, and the presence of external drivers (e.g., weather, tariff changes). This heterogeneity induces strong distribution shifts and non‑stationarities across the successive tasks, thereby rendering the continual‑learning problem genuinely demanding even though the downstream objective (forecasting a scalar energy quantity) is homogeneous. 
We argue that this configuration mirrors real‑world operational settings for several reasons: utilities constantly add or retire sensors and renewable assets, producing new temporal patterns; regulatory or market shifts (e.g., tariff changes, demand‑response schemes) repeatedly alter consumption dynamics; and privacy constraints mean many relevant series remain unseen during pre‑training, so models must continually adapt while retaining prior knowledge.


\textbf{$\dataEnedisResidentiel$} \footnote{\url{https://opendata.enedis.fr/datasets/bilan-electrique-jour}} is a daily dataset, covering five years of residential electricity consumption from 2021 to 2025. It corresponds to the variable “consommation télérélevée résidentielle”, representing the average daily power consumption for low-voltage residential users (in watts).

\textbf{$\dataEnedisPhoto$} is a daily dataset from the open data repository, covering five years of photovoltaic production from 2021 to 2025. It corresponds to the variable “production\_photovoltaique”, representing the average daily photovoltaic power production (in watts).

\textbf{$\dataElisa$} 
\footnote{\url{https://zenodo.org/records/17199246}}
introduces a single daily private household of energy consumption over two years, exposing the model to long-term temporal dynamics, including yearly seasonality and potential behavioral drift, representing a rare and unseen entity during the pretraining phase of TSFMs. 

\textbf{$\dataEnedisEolien$} is a daily dataset from the open data repository, covering five years of wind power production from 2021 to 2025. It corresponds to the variable “production\_eolien”, representing the average daily total wind power production (in watts).


\subsection{Experimental Setup}

All models were evaluated under the same forecasting protocol. Each time series was transformed into supervised samples using a sliding-window strategy with stride 1, using a context window of length $L=256$ and a forecasting horizon of length $H=128$. The resulting windows were chronologically split into training, validation, and test sets using a 70\%/15\%/15\% ratio.

To account for scale differences across datasets, we adopted per-window normalization: each context window was normalized using its own mean and standard deviation, and the corresponding forecasting horizon was normalized using the same statistics.

SamFormer was trained with Adam and SAM, while TSFMs were fine-tuned with Adam. A batch size of 64 was used for all experiments, and learning rates were selected through grid search over $\{10^{-3}, 10^{-4}, 10^{-5}\}$.

For each model and task, we first determined the optimal number of epochs using early stopping on the validation loss (patience of 10 epochs, maximum 100 epochs). The selected epoch budgets were then fixed and reused in all continual learning experiments, including DER, following recommendations from the continual learning literature \cite{der2020}, to ensure fair comparisons across methods.
\subsection{Evaluation protocol}
\label{subsection:Evaluation-protocol}
To systematically evaluate continual learning methods on time series forecasting, we define four learning protocols. Given a scenario $\mathcal{S} = (\mathcal{D}_1, \ldots, \mathcal{D}_N)$, they are defined as follows:


\paragraph{Sequential training} is the main setting described so far. It consists in successively training on $\mathcal{D}_1$, then $\mathcal{D}_2$, \ldots{}, until $\mathcal{D}_N$. It leads to $N$ training stages, with no preserved training data from one dataset to another. At stage $t$, training is performed on $\mathcal{D}_t$ only.

\paragraph{Cumulative training}
corresponds to a continual setting in which, at stage $t$, the model has access to whole datasets it has seen so far. It simulates an infinite memory buffer as a kind of upper bound regarding sequential training.
At stage $t$, training is performed on $\{\mathcal{D}_1, \ldots{}, \mathcal{D}_t\}$.

\paragraph{Independent training}
consists in performing $N$ independent trainings. Each training is dedicated to a single dataset. It is used as a non-continual baseline to estimate the difficulty of each dataset and to highlight any cross-dataset behaviour; for instance, two datasets may help each other due to similarities.

\paragraph{Offline training}
is another non-continual baseline for which a single training is performed over all the datasets of a given scenario at once. It assumes all data is available at start. The idea here is to highlight the best we could reach, taking into account the model ability to model all datasets' distributions and specificities.









To evaluate the forecasting performance of the model, we rely on the root mean square error (RMSE) metric, computed between the prediction $\hat{y}$ and ground truth $y$ as follows:
\begin{equation}
\text{RMSE}(\hat{y}, y)=
\sqrt{
\frac{1}{H}
\sum_{t=1}^{H}
(\hat{y}_t-y_t)^2
},
\end{equation}
with $H$ the forecast horizon.
For continual learning settings, we note $\text{RMSE}^{(j\rightarrow i)}$ the RMSE associated to the evaluation on dataset $\mathcal{D}_i$, after the model was fine-tuned on $\mathcal{D}_j$.

To assess the catastrophic forgetting phenomenon, we rely on the backward transfer (BWT) metric. Regarding dataset $\mathcal{D}_i$, it compares the final performance on $\mathcal{D}_i$ with the performance obtained immediately after training on $\mathcal{D}_i$:

\begin{equation}
\text{BWT}^{(i)} = \text{RMSE}^{(N \rightarrow i)} - \text{RMSE}^{(i \rightarrow i)}.
\end{equation}

In this way, the more negative the backward transfer, the greater the forgetting.

\subsection{Measuring the inter-dataset transferability}
To design a relevant continual learning scenario, it is important to ensure that catastrophic forgetting emerges during training. To do so, it is necessary to quantify how training on one dataset impacts the performance on another dataset of the same scenario. 
We noted that standard similarity metrics such as Dynamic Time Warping (DTW) fail to capture such characteristics. Indeed, transferability is intrinsically related to a given model and its way of extracting features. This is discussed in Section \ref{sec:expe-cst}, motivating the design of a model-aware metric.

We propose a novel metric, the cross-series transferability (CST), to quantify the knowledge transferability of one time-series dataset to another. To estimate how knowledge from dataset $\mathcal{D}_i$ can be transferred to dataset $\mathcal{D}_j$, this model-dependent metric is defined as a performance ratio on $\mathcal{D}_j$ between the two configurations: after training on $\mathcal{D}_i$ and from random initialization. The goal here is to quantify the positive contribution of training on $\mathcal{D}_i$ compared to an initial state $\varnothing{}$, whether it is from scratch ("no knowledge", random weights) or from a pretrained model (TSFMs).

\begin{equation}
    \text{CST}^{(i \rightarrow j)} = \frac{RMSE^{(i \rightarrow j)}}{RMSE^{(\varnothing{} \rightarrow j)}},
\end{equation}
where $RMSE^{(\varnothing{} \rightarrow j)}$ represents the evaluation of the model in its initial state (random for Samformer, or pretrained weights for TSFMs) over $\mathcal{D}_j$.

As one can note, this definition is asymetric, meaning that training on $\mathcal{D}_i$ can be more useful to forecast on $\mathcal{D}_j$ than training on $\mathcal{D}_j$ to forecast on $\mathcal{D}_i$. Also, there is no constraint on CST values that can be either lower than 1, highlighting a good transferability, or greater than 1, reflecting an opposite knowledge by reducing performance.


\section{Results}
\label{sec:expe}
First, we present the CST metric applied on Samformer and how to use it to anticipate catastrophic forgetting. Then, we compare the various TSFM models (TimesFM 2.0, Chronos-base, and Chronos-small) with one another, and against the specific small model SAMformer over both defined scenarios and the four learning protocols. Finally, we study the impact of Dark Experience Replay on the models' sensitivity to catastrophic forgetting.

All results, for the synthetic scenario \scenarioSynthetic{} and the real-world scenario \scenarioReal{}, are shown in Table \ref{tabl:synthCL} and \ref{tabl:realCL}, respectively.

\subsection{Designing continual learning scenario for time series}
\label{sec:expe-cst}


\begin{figure}[htbp!]
    \centering
    \includegraphics[width=0.45\linewidth]{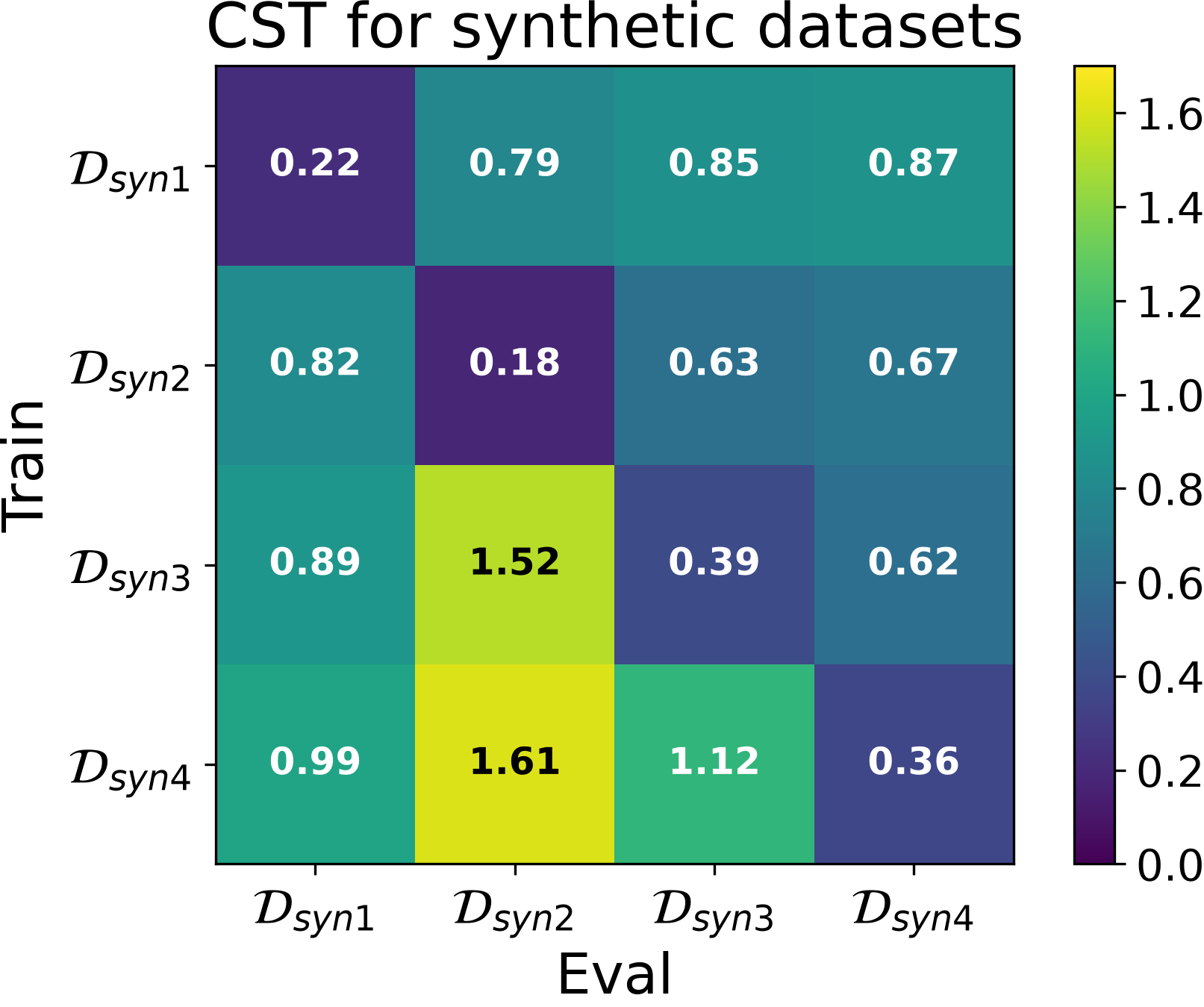}
    \hfill
    \includegraphics[width=0.45\linewidth]{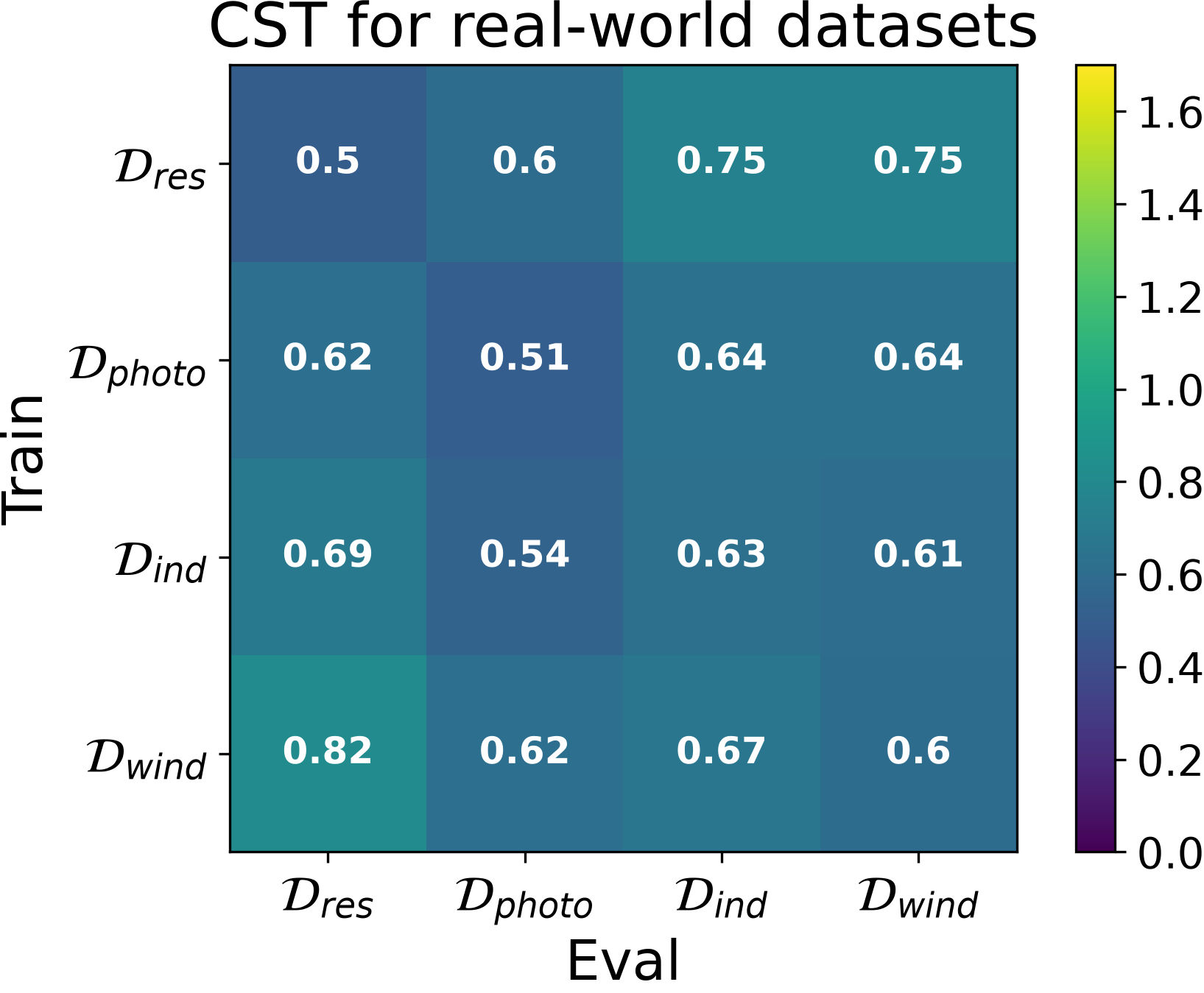}
    \caption{CST matrices for synthetic datasets (left) and real-world datasets (right) regarding Samformer architecture. The lower the values, the more the training dataset facilitates forecasting on the evaluation dataset.}
    \label{fig:cst}
\end{figure}

Figure \ref{fig:cst} provides the CST matrices for synthetic and real-world scenarios for the Samformer model. First, one can note that, for both scenarios, the minimum value of each column is always on the diagonal of the matrix. This means that the training set of the target dataset ("Eval") is always the one that best transfers to the test set of the target dataset. 
A second observation is that the values outside the diagonal are, on average, lower for the real-word datasets than for the synthetic datasets. They are also always below 1. This means that there are more knowledge transfer opportunities for the real-world datasets, and thus less catastrophic forgetting. This is confirmed by the average BWT of the sequential training section which is higher for \scenarioSynthetic{} (2.04) than for \scenarioReal{} (0.45) in Tables \ref{tabl:synthCL} and \ref{tabl:realCL}.
A third point is about the values greater than one (\scenarioSynthetic{}). These values directly correlate with the most severe forgetting situations. Indeed, the $\text{CST}^{\dataSynC \rightarrow \dataSynB} = 1.52$ and corresponds to an increase of RMSE of 2.56 points over \dataSynB{} when switching from training on \dataSynB (0.33) to training on \dataSynC (2.86). Similarly, $\text{CST}^{\dataSynD \rightarrow \dataSynC} = 1.12$ results in an increase of 1.65 points (from 1.01 to 2.66).
In this way, computing these matrices enables the anticipation of the forgetting property when switching from one dataset to another.

As mentioned previously, such estimation cannot be performed by the standard DTW metric. Indeed, experiments on \scenarioReal{} revealed that DTW identifies $\dataEnedisResidentiel$ and $\dataElisa$ as highly similar (distance 0.037), while $\dataEnedisPhoto$ is found to be closer to $\dataEnedisEolien$ (0.322) than to $\dataElisa$ (0.395). However, the forgetting severity observed during sequential learning does not follow this ordering. This discrepancy suggests that catastrophic forgetting depends not only on similarities in the observation space but also on the internal representations learned by the forecasting model.

\subsection{Evaluation of catastrophic forgetting}

In this part, we focus on the first four learning  protocols presented in section \ref{subsection:Evaluation-protocol}, while the results of DER are discussed separately in section \ref{subsection: DER}.

\begin{table*}[t!]
\centering
\scriptsize
\setlength{\tabcolsep}{2pt}
\renewcommand{\arraystretch}{0.88}

\resizebox{\textwidth}{!}{
\begin{tabular}{l|c c c c c|c c c c c}
\hline
Models
& \multicolumn{5}{c|}{\textbf{Samformer (100k)}} 
& \multicolumn{5}{c}{\textbf{TimesFM 2.0 (500M)}} \\

Datasets 
& \dataSynA & \dataSynB & \dataSynC & \dataSynD & Mean 
& \dataSynA & \dataSynB & \dataSynC & \dataSynD & Mean  \\
\hline

\tabSectionName{Independent training} 
& 0.45 & 0.34 & 1.00 & 1.17 & 0.74 
& 0.11 & 0.02 & 1.31 & 0.82 & \textbf{0.57} \\

\hline

\tabSectionName{Offline training} 
& 0.89 & 0.86 & 1.50 & 1.95 & 1.30
& 0.13 & 0.04 & 1.57 & 1.01 & \textbf{0.70}\\

\hline
\tabSectionName{Cumulative training} & & & & & & & & & & \\

\dataSynA 
& 0.45 
& \textcolor{futuretask}{1.59} 
& \textcolor{futuretask}{2.11} 
& \textcolor{futuretask}{2.74}
&
& 0.11 
& \textcolor{futuretask}{0.99} 
& \textcolor{futuretask}{1.39} 
& \textcolor{futuretask}{1.75}
&\\

\dataSynA+\dataSynB
& 0.56 & 0.50 
& \textcolor{futuretask}{2.22} 
& \textcolor{futuretask}{2.44}
&

& 0.12 & 0.05 
& \textcolor{futuretask}{1.36} 
& \textcolor{futuretask}{1.67}
&\\

\dataSynA+\dataSynB+\dataSynC
& 0.85 & 0.70 & 1.52 
& \textcolor{futuretask}{2.08}
&

& 0.10 & 0.09 & 1.37 
& \textcolor{futuretask}{1.57}
&\\

\dataSynA+\dataSynB+\dataSynC+\dataSynD 
& 0.95 & 0.81 & 1.47 & 1.99 & 1.31
& 0.10 & 0.03 & 1.44 & 0.77 & \textbf{0.58} \\

\cline{2-11}
Average BWT ($\downarrow$)
& \multicolumn{5}{c|}{0.25}
& \multicolumn{5}{c}{0.01}\\

\hline

\tabSectionName{Sequential training} & & & & & & & & & & \\

after \dataSynA 
& 0.45 
& \textcolor{futuretask}{1.59} 
& \textcolor{futuretask}{2.11} 
& \textcolor{futuretask}{2.74}
& 
&

0.11 
& \textcolor{futuretask}{0.99} 
& \textcolor{futuretask}{1.39} 
& \textcolor{futuretask}{1.75}
& \\

after \dataSynB 
& 1.65 & 0.33 
& \textcolor{futuretask}{1.60} 
& \textcolor{futuretask}{2.08}
&

& 0.47 & 0.05 
& \textcolor{futuretask}{1.30} 
& \textcolor{futuretask}{1.49}
&\\

after \dataSynC 
& 1.86 & 2.86 & 1.01 
& \textcolor{futuretask}{1.88}
&

& 0.46 & 0.85 & 1.28 
& \textcolor{futuretask}{1.55}
&\\

after \dataSynD 
& 2.04 & 3.22 & 2.66 & 1.20 & 2.28
& 0.68 & 1.46 & 1.59 & 0.77 & \textbf{1.13}\\

\cline{2-11}
Average BWT ($\downarrow$)
& \multicolumn{5}{c|}{2.04}
& \multicolumn{5}{c}{\textbf{0.77}}\\

\hline
\hline
\tabSectionName{Dark Experience Replay} & & & & & & & & & \\

after \dataSynA
& 0.45 & \textcolor{futuretask}{1.59} & \textcolor{futuretask}{2.11} & \textcolor{futuretask}{2.74} & 
& 0.11 & \textcolor{futuretask}{0.99} & \textcolor{futuretask}{1.39} & \textcolor{futuretask}{1.75} & \\

after \dataSynB
& 0.43 & 0.54 & \textcolor{futuretask}{2.49} & \textcolor{futuretask}{2.57} & 
& 0.13 & 0.05 & \textcolor{futuretask}{1.34} & \textcolor{futuretask}{1.64} & \\

after \dataSynC
& 0.66 & 0.60 & 1.63 & \textcolor{futuretask}{2.28} & 
& 0.13 & 0.14 & 1.38 & \textcolor{futuretask}{1.52} & \\

after \dataSynD
& 0.67 & 0.65 & 1.75 & 2.09 & 1.29 
& 0.17 & 0.18 & 1.43 & 0.81 & \textbf{0.65}\\

\cline{2-11}
Average BWT ($\downarrow$)
& \multicolumn{5}{c|}{0.15}
& \multicolumn{5}{c}{0.08}\\
\hline
\end{tabular}
}

\vspace{0.75cm}

\resizebox{\textwidth}{!}{
\begin{tabular}{l|c c c c c|c c c c c}
\hline
Models
& \multicolumn{5}{c|}{\textbf{Chronos-base (120M)}} 
& \multicolumn{5}{c}{\textbf{Chronos-small (28M)}} \\

Datasets 
& \dataSynA & \dataSynB & \dataSynC & \dataSynD & Mean 
& \dataSynA & \dataSynB & \dataSynC & \dataSynD & Mean  \\
\hline

\tabSectionName{Independent training} 
& 0.12 & 0.02 & 1.64 & 0.69 & 0.62 
& 0.13 & 0.05 & 1.51 & 0.99 & 0.67 \\

\hline

\tabSectionName{Offline training} 
& 0.24 & 0.02 & 1.76 & 1.20 & 0.81
& 0.16 & 0.05 & 1.76 & 0.97 & 0.74 \\

\hline
\tabSectionName{Cumulative training} & & & & & & & & & & \\

\dataSynA 
& 0.12 
& \textcolor{futuretask}{0.50} 
& \textcolor{futuretask}{1.59} 
& \textcolor{futuretask}{1.81}
&
& 0.13 
& \textcolor{futuretask}{1.12} 
& \textcolor{futuretask}{1.61} 
& \textcolor{futuretask}{2.00}
& \\

\dataSynA+\dataSynB
& 0.19 & 0.03 
& \textcolor{futuretask}{1.75} 
& \textcolor{futuretask}{1.93}
&

& 0.14 & 0.04 
& \textcolor{futuretask}{1.83} 
& \textcolor{futuretask}{1.92} 
& \\

\dataSynA+\dataSynB+\dataSynC
& 0.22 & 0.01 & 1.66 
& \textcolor{futuretask}{1.97}
&

& 0.13 & 0.05 & 1.57 
& \textcolor{futuretask}{1.87} 
& \\

\dataSynA+\dataSynB+\dataSynC+\dataSynD 
& 0.22 & 0.02 & 1.85 & 1.45 & 0.89
& 0.14 & 0.04 & 1.58 & 0.69 & 0.61\\

\cline{2-11}
Average BWT ($\downarrow$)
& \multicolumn{5}{c|}{0.09}
& \multicolumn{5}{c}{\textbf{0.004}} \\

\hline
\tabSectionName{Sequential training} & & & & & & & & & \\

after \dataSynA 
& 0.12 
& \textcolor{futuretask}{0.50} 
& \textcolor{futuretask}{1.59} 
& \textcolor{futuretask}{1.81}
&
&

0.13 
& \textcolor{futuretask}{1.12} 
& \textcolor{futuretask}{1.61} 
& \textcolor{futuretask}{2.00}
& \\

after \dataSynB 
& 0.99 & 0.02 
& \textcolor{futuretask}{1.61} 
& \textcolor{futuretask}{1.83}
&

& 0.99 & 0.04 
& \textcolor{futuretask}{1.78} 
& \textcolor{futuretask}{1.81} 
& \\

after \dataSynC 
& 1.48 & 0.74 & 1.59 
& \textcolor{futuretask}{1.93}
&

& 1.12 & 0.87 & 1.52 
& \textcolor{futuretask}{1.92} 
& \\

after \dataSynD 
& 2.55 & 1.89 & 1.91 & 0.71 & 1.77
& 1.57 & 1.17 & 1.74 & 0.65 & 1.28 \\

\cline{2-11}
Average BWT ($\downarrow$)
& \multicolumn{5}{c|}{1.54}
& \multicolumn{5}{c}{0.93} \\

\hline
\hline
\tabSectionName{Dark Experience Replay} & & & & & & & & & \\

after \dataSynA
& 0.12 & \textcolor{futuretask}{0.50} & \textcolor{futuretask}{1.59} & \textcolor{futuretask}{1.81} &
& 0.13 & \textcolor{futuretask}{1.12} & \textcolor{futuretask}{1.61} & \textcolor{futuretask}{2.00} & \\

after \dataSynB
& 0.21 & 0.03 & \textcolor{futuretask}{1.66} & \textcolor{futuretask}{1.84} &
& 0.16 & 0.10 & \textcolor{futuretask}{1.91} & \textcolor{futuretask}{1.86} & \\

after \dataSynC
& 0.26 & 0.04 & 1.62 & \textcolor{futuretask}{1.95} &
& 0.18 & 0.08 & 1.58 & \textcolor{futuretask}{1.82} & \\

after \dataSynD
& 0.25 & 0.05 & 1.50 & 1.18 & 0.75
& 0.23 & 0.10 & 1.68 & 0.99 & 0.75\\

\cline{2-11}
Average BWT ($\downarrow$)
& \multicolumn{5}{c|}{\textbf{0.01}}
& \multicolumn{5}{c}{0.07} \\

\hline
\end{tabular}
}

\caption{RMSE performance comparison of TSFMs and Samformer on synthetic scenario \scenarioSynthetic{} under independent, offline, cumulative, sequential and DER training strategies. Best performance per setting is shown in bold; gray indicates zero-shot evaluation.}
\label{tabl:synthCL}
\end{table*}

\begin{table*}[htbp!]
\centering
\scriptsize
\setlength{\tabcolsep}{2.5pt}
\renewcommand{\arraystretch}{0.92}

\resizebox{\textwidth}{!}{
\begin{tabular}{l|c c c c c|c c c c c}
\hline
Models
& \multicolumn{5}{c|}{\textbf{Samformer (100k)}}
& \multicolumn{5}{c}{\textbf{TimesFM 2.0 (500M)}} \\

Datasets
& \dataEnedisResidentiel & \dataEnedisPhoto & \dataElisa & \dataEnedisEolien & Mean
& \dataEnedisResidentiel & \dataEnedisPhoto & \dataElisa & \dataEnedisEolien & Mean\\
\hline

\tabSectionName{Independent training}
& 1.66 & 1.09 & 1.43 & 1.23 & \textbf{1.35}
& 1.82 & 1.12 & 1.69 & 1.26 & 1.47\\

\hline

\tabSectionName{Offline training}
& 1.87 & 1.13 & 1.47 & 1.31 & 1.44
& 1.67 & 1.17 & 1.40 & 1.30 & \textbf{1.39}\\

\hline

\tabSectionName{Cumulative training} & & & & & & & & & \\

\dataEnedisResidentiel
& 1.66 
& \textcolor{futuretask}{1.29} 
& \textcolor{futuretask}{1.69} 
& \textcolor{futuretask}{1.53}
&
& 1.82 
& \textcolor{futuretask}{1.22} 
& \textcolor{futuretask}{1.72} 
& \textcolor{futuretask}{1.36} \\

\dataEnedisResidentiel + \dataEnedisPhoto
& 1.73 & 1.18 
& \textcolor{futuretask}{1.58} 
& \textcolor{futuretask}{1.41}
&
& 1.71 & 1.11 
& \textcolor{futuretask}{1.61} 
& \textcolor{futuretask}{1.66} \\

\dataEnedisResidentiel + \dataEnedisPhoto + \dataElisa
& 1.80 & 1.14 & 1.51 
& \textcolor{futuretask}{1.36}
&
& 1.79 & 1.19 & 1.56 
& \textcolor{futuretask}{1.42} \\

\dataEnedisResidentiel + \dataEnedisPhoto + \dataElisa + \dataEnedisEolien
& 1.90 & 1.12 & 1.46 & 1.31 & \textbf{1.45}
& 1.89 & 1.22 & 1.54 & 1.32 & 1.49\\

\cline{2-11}
Average BWT ($\downarrow$)
& \multicolumn{5}{c|}{0.04}
& \multicolumn{5}{c}{0.05} \\

\hline

\tabSectionName{Sequential training} & & & & & & & & &  \\

after \dataEnedisResidentiel
& 1.66 
& \textcolor{futuretask}{1.29} 
& \textcolor{futuretask}{1.69} 
& \textcolor{futuretask}{1.53}
&
& 1.82 
& \textcolor{futuretask}{1.22} 
& \textcolor{futuretask}{1.72} 
& \textcolor{futuretask}{1.36} \\

after \dataEnedisPhoto
& 2.01 & 1.10 
& \textcolor{futuretask}{1.45} 
& \textcolor{futuretask}{1.32}
&
& 1.84 & 1.11 
& \textcolor{futuretask}{1.51} 
& \textcolor{futuretask}{1.71} \\

after \dataElisa
& 2.25 & 1.15 & 1.43 
& \textcolor{futuretask}{1.25}
&
& 2.07 & 1.34 & 1.66 
& \textcolor{futuretask}{1.42} \\

after \dataEnedisEolien
& 2.69 & 1.33 & 1.52 & 1.23 & 1.69
& 2.47 & 1.33 & 1.55 & 1.25 & \textbf{1.65} \\

\cline{2-11}
Average BWT ($\downarrow$)
& \multicolumn{5}{c|}{ 0.45}
& \multicolumn{5}{c}{\textbf{0.25}} \\

\hline

\tabSectionName{Dark Experience Replay} & & & & & & & & & \\

after \dataEnedisResidentiel
& 1.66 & \textcolor{futuretask}{1.29} & \textcolor{futuretask}{1.69} & \textcolor{futuretask}{1.53} & 
& 1.82 & \textcolor{futuretask}{1.22} & \textcolor{futuretask}{1.72} & \textcolor{futuretask}{1.36} & \\

after \dataEnedisPhoto
& 1.70 & 1.22 & \textcolor{futuretask}{1.60} & \textcolor{futuretask}{1.45} & 
& 1.67 & 1.13 & \textcolor{futuretask}{1.60} & \textcolor{futuretask}{1.63} & \\

after \dataElisa
& 1.74 & 1.19 & 1.56 & \textcolor{futuretask}{1.41} & 
& 1.98 & 1.25 & 1.55 & \textcolor{futuretask}{1.52} & \\

after \dataEnedisEolien
& 1.75 & 1.19 & 1.54 & 1.38 & 1.47
& 1.65 & 1.30 & 1.47 & 1.31 & \textbf{1.43}\\

\cline{2-11}
Average BWT ($\downarrow$)
& \multicolumn{5}{c|}{0.02}
& \multicolumn{5}{c}{-0.03} \\

\hline
\end{tabular}
}

\vspace{0.75cm}

\resizebox{\textwidth}{!}{
\begin{tabular}{l|c c c c c|c c c c c}
\hline
Models
& \multicolumn{5}{c|}{\textbf{Chronos-base (120M)}}
& \multicolumn{5}{c}{\textbf{Chronos-small (28M)}} \\

Datasets
& \dataEnedisResidentiel & \dataEnedisPhoto & \dataElisa & \dataEnedisEolien & Mean
& \dataEnedisResidentiel & \dataEnedisPhoto & \dataElisa & \dataEnedisEolien & Mean \\
\hline

\tabSectionName{Independent training}
& 1.83 & 1.05 & 1.44 & 1.35 & 1.42
& 1.87 & 1.10 & 1.44 & 1.35 & 1.44\\

\hline

\tabSectionName{Offline training}
& 1.76 & 1.19 & 1.92 & 1.38 & 1.56
& 1.84 & 1.24 & 1.69 & 1.39 & 1.54\\

\hline 
\tabSectionName{Cumulative training} & & & & & & & & & \\

\dataEnedisResidentiel
& 1.83 
& \textcolor{futuretask}{1.46} 
& \textcolor{futuretask}{1.65} 
& \textcolor{futuretask}{1.35}
&
& 1.87 
& \textcolor{futuretask}{1.52} 
& \textcolor{futuretask}{1.93} 
& \textcolor{futuretask}{1.34} \\

\dataEnedisResidentiel + \dataEnedisPhoto
& 1.80 & 1.12 
& \textcolor{futuretask}{1.51} 
& \textcolor{futuretask}{1.30}
&
& 1.72 & 1.17 
& \textcolor{futuretask}{1.50} 
& \textcolor{futuretask}{1.42} \\

\dataEnedisResidentiel + \dataEnedisPhoto + \dataElisa
& 1.84 & 1.16 & 1.58 
& \textcolor{futuretask}{1.57}
&
& 1.96 & 1.14 & 1.61 
& \textcolor{futuretask}{1.49} \\

\dataEnedisResidentiel + \dataEnedisPhoto + \dataElisa + \dataEnedisEolien
& 1.77 & 1.21 & 1.56 & 1.40 & 1.48
& 1.72 & 1.19 & 1.76 & 1.35 & 1.50 \\

\cline{2-11}
Average BWT ($\downarrow$)
& \multicolumn{5}{c|}{\textbf{0.00}}
& \multicolumn{5}{c}{\textbf{0.00}} \\

\hline

\tabSectionName{Sequential training} & & & & & & & & & \\

after \dataEnedisResidentiel
& 1.83 
& \textcolor{futuretask}{1.46} 
& \textcolor{futuretask}{1.65} 
& \textcolor{futuretask}{1.32}
&
& 1.87 
& \textcolor{futuretask}{1.52} 
& \textcolor{futuretask}{1.93} 
& \textcolor{futuretask}{1.34} \\

after \dataEnedisPhoto
& 1.67 & 1.08 
& \textcolor{futuretask}{1.48} 
& \textcolor{futuretask}{1.28}
&
& 1.93 & 1.11 
& \textcolor{futuretask}{1.61} 
& \textcolor{futuretask}{1.64} \\

after \dataElisa
& 2.09 & 1.17 & 1.46 
& \textcolor{futuretask}{1.35}
&
& 2.12 & 1.16 & 1.46 
& \textcolor{futuretask}{1.34} \\

after \dataEnedisEolien
& 2.48 & 1.27 & 1.60 & 1.27 & 1.66
& 2.82 & 1.38 & 1.82 & 1.34 & 1.83\\

\cline{2-11}
Average BWT ($\downarrow$)
& \multicolumn{5}{c|}{0.33}
& \multicolumn{5}{c}{0.52} \\

\hline

\tabSectionName{Dark Experience Replay} & & & & & & & & & \\

after \dataEnedisResidentiel
& 1.83 & \textcolor{futuretask}{1.46} & \textcolor{futuretask}{1.65} & \textcolor{futuretask}{1.32} &
& 1.87 & \textcolor{futuretask}{1.52} & \textcolor{futuretask}{1.93} & \textcolor{futuretask}{1.34} & \\

after \dataEnedisPhoto
& 1.96 & 1.32 & \textcolor{futuretask}{1.58} & \textcolor{futuretask}{1.27} &
& 1.71 & 1.28 & \textcolor{futuretask}{1.56} & \textcolor{futuretask}{1.32} & \\

after \dataElisa
& 2.05 & 1.15 & 1.50 & \textcolor{futuretask}{1.26} & 
& 1.82 & 1.16 & 1.50 & \textcolor{futuretask}{1.30} & \\

after \dataEnedisEolien
& 1.86 & 1.15 & 1.50 & 1.26 & 1.44
& 1.79 & 1.22 & 1.53 & 1.23 & 1.44\\

\cline{2-11}
Average BWT ($\downarrow$) 
& \multicolumn{5}{c|}{\textbf{-0.05}}
& \multicolumn{5}{c}{-0.04} \\
\hline
\end{tabular}
}

\caption{RMSE performance comparison of TSFMs and Samformer on real-world scenario \scenarioReal{} under independent, offline, cumulative, sequential and DER training strategies. Best performance per setting is shown in bold; gray indicates zero-shot evaluation.}
\label{tabl:realCL}
\end{table*}

Focusing on the independent training section, one can note that the order of dataset difficulty (\dataEnedisPhoto{}, \dataEnedisEolien{}, \dataElisa, \dataEnedisResidentiel{}) for \scenarioReal{} with increasing RMSE, is shared for all models. For \scenarioSynthetic{}, the order is also rather stable with an inversion of \dataSynC{} and \dataSynD{} as the more difficult dataset between all TSFMs on one side and Samformer on the other. We assume that this is due to the pretraining data of TSFMs that could be closer to \dataSynD{} than to \dataSynC{}, or to the increased expressiveness of TSFM models due to their higher number of parameters.

Regarding sequential training section, it should be noted that all models exhibit a clear catastrophic forgetting behavior. This can be observed by comparing the diagonal values to the bottom line of the section ("after \dataSynD{}" for \scenarioSynthetic{} or "after \dataEnedisEolien{}" for \scenarioReal{}). For instance, considering Samformer on \scenarioSynthetic{}, the RMSE associated to \dataSynA{} increases from 0.45 to 2.04. The same goes for \dataSynB{} with an RMSE going from 0.33 to 3.22, and for \dataSynC{} whose RMSE increases from 1.01 up to 2.33. This catastrophic forgetting is also noticeable through the average BWT metric. In this way, one can note that Samformer is the most sensitive approach for \scenarioSynthetic{} with a BWT of 2.04, but it is Chronos-small for \scenarioReal{} with 0.52 of BWT.
For both scenarios, TimesFM 2.0 reaches the best average RMSE and BWT on that setting with 1.13 and 0.77 for \scenarioSynthetic{}, and 1.65 and 0.25 for \scenarioReal{}. More generally, this experiment suggests that a specific model trained from scratch (Samformer) is not a good option for this naive continual learning setting, as it is the worst option for \scenarioSynthetic{} and the second worst for \scenarioReal{}.
Interestingly, the catastrophic forgetting is more severe for the synthetic scenario than for the real-world scenario. This suggests that, despite being noisier and harder to forecast, the real-world datasets remain more semantically related than the synthetic tasks, which were intentionally designed with strongly different spectral compositions. In particular, all real-world datasets belong to the same broad energy-consumption domain and therefore likely share partially compatible temporal structures and latent representations. As a result, sequential adaptation induces weaker representation mismatch than in the synthetic setting, where moving from \dataSynA/\dataSynB  toward \dataSynC/\dataSynD~introduced abrupt spectral shifts.

The cumulative training section highlights that the performance could be better for all models: the limitation is not about the model capabilities but the training strategy. Indeed, the average RMSE and BWT are lower for cumulative training setting than for sequential training setting, for both scenarios and for all models.

The offline training setting can be viewed as a naive lower bound: we could not expect the model to perform better on all datasets compared to sequential training. Interestingly, however, the results demonstrate that cumulative training can lead to better results than offline training. For instance, offline training leads to an average RMSE of 0.70 for TimesFM 2.0 on \scenarioSynthetic{}, but it decreases to 0.58 for cumulative training. This can be thought as a curriculum training effect, making training of complex data more efficient with progressive complexity.

\subsection{Impact of Dark Experience Replay}
\label{subsection: DER}
The last section of Tables \ref{tabl:synthCL} and \ref{tabl:realCL} provide the results of the DER strategy for all models on scenario \scenarioSynthetic{} and \scenarioReal{}, respectively.

The Dark Experience Replay strategy clearly mitigates the catastrophic forgetting phenomenon. When comparing naive sequential training to DER, one can observe that the average RMSE metric decreases in all cases (from 2.28 to 1.19 for Samformer on \scenarioSynthetic{} or from 1.83 to 1.44 for Chronos-small on \scenarioReal{}). The same goes for the BWT. The most noticeable case is Samformer on \scenarioSynthetic{} for which the BWT decreases from 2.04 down to 0.15. Even the smallest improvement is significant (from 0.25 to -0.03 for TimesFM 2.0 on \scenarioReal{}). Furthermore, DER even succeeds in reaching negative BWT values, such as this latter value, reflecting not only the disappearance of forgetting (on average), but also the positive interaction between the different datasets.
A concrete example is for Chronos-base on \scenarioReal{}, which reaches an average RMSE of 1.44 for DER strategy, which is lower than that for cumulative training (1.48) and offline training (1.56).

\section{Conclusion}
\label{sec:conclu}
In this work, we conducted an extensive continual learning analysis comparing a small task-specific forecasting model and several time-series foundation models (TSFMs) under both synthetic and real-world sequential forecasting scenarios. 
Our experiments reveal that continual learning behavior strongly depends not only on forecasting difficulty, but also on the compatibility of latent temporal representations across datasets. The proposed model-dependent CST metric successfully anticipates the severity of the forgetting phenomenon when switching training from one dataset to another.
On synthetic datasets, although the signals are relatively easy to predict due to their sinusoidal structure, the tasks become increasingly different in representation space from \dataSynA{}/\dataSynB{} to \dataSynC{}/\dataSynD{}. This strong spectral shift leads to severe catastrophic forgetting during naive sequential fine-tuning, especially when moving from simpler harmonic patterns toward denser multi-frequency regimes. Replay-based rehearsal with DER proves highly effective at mitigating forgetting across all architectures, substantially reducing both RMSE and BWT. Although TSFMs are better suited for the complex synthetic scenario, we demonstrate that a very small model (Samformer) trained from scratch can be competitive with TSFMs on a real-world scenario with DER strategy.



\bibliographystyle{splncs04}
\bibliography{these_nouha}

@article{reservoir,
author = {Vitter, Jeffrey S.},
title = {Random sampling with a reservoir},
year = {1985},
issue_date = {March 1985},
publisher = {Association for Computing Machinery},
address = {New York, NY, USA},
volume = {11},
number = {1},
journal = {ACM Trans. Math. Softw.},
pages = {37–57},
numpages = {21}
}

@article{radford2018improving,
  title={Improving language understanding by generative pre-training},
  author={Radford, Alec and Narasimhan, Karthik and Salimans, Tim and Sutskever, Ilya and others},
  year={2018},
  publisher={openAI}
}

@inproceedings{devlin2019bert,
  title={Bert: Pre-training of deep bidirectional transformers for language understanding},
  author={Devlin, Jacob and Chang, Ming-Wei and Lee, Kenton and Toutanova, Kristina},
  booktitle={Proceedings of the 2019 conference of the North American chapter of the association for computational linguistics: human language technologies, vol 1},
  pages={4171--4186},
  year={2019}
}

@article{kirkpatrick2017overcoming,
  title={Overcoming catastrophic forgetting in neural networks},
  author={Kirkpatrick, James and et al.},
  journal={Proceedings of the national academy of sciences},
  volume={114},
  number={13},
  pages={3521--3526},
  year={2017},
  publisher={National Academy of Sciences}
}

@inproceedings{der2020,
author = {Buzzega, Pietro and Boschini, Matteo and Porrello, Angelo and Abati, Davide and Calderara, Simone},
title = {Dark experience for general continual learning: a strong, simple baseline},
year = {2020},
isbn = {9781713829546},
publisher = {Curran Associates Inc.},
address = {Red Hook, NY, USA},
booktitle = {Proceedings of the 34th International Conference on Neural Information Processing Systems},
articleno = {1335},
numpages = {11},
location = {Vancouver, BC, Canada},
series = {NIPS '20}
}

@inproceedings{zhou2021informer,
  title={Informer: Beyond efficient transformer for long sequence time-series forecasting},
  author={Zhou, Haoyi and et al.},
  booktitle={Proceedings of the AAAI conference on artificial intelligence},
  volume={35},
  number={12},
  pages={11106--11115},
  year={2021}
}

@article{bahrpeyma2021methodology,
  title={A methodology for validating diversity in synthetic time series generation},
  author={Bahrpeyma, Fouad and Roantree, Mark and Cappellari, Paolo and Scriney, Michael and McCarren, Andrew},
  journal={MethodsX},
  volume={8},
  pages={101459},
  year={2021},
  publisher={Elsevier}
}

@inproceedings{radford2021learning,
  title={Learning Transferable Visual Models From Natural Language Supervision},
  author={Radford, Alec and  et al.},
  booktitle={International Conference on Machine Learning (ICML)},
  year={2021}
}

@article{van2019three,
  title={Three types of incremental learning},
  author={van de Ven, G.M. and Tuytelaars, T. and Tolias, A.S},
  journal={Nature Mach Intell 4,},
 pages={1185–1197},
  year={2022}
}

@inproceedings{matteoni2022continual,
  author       = {Federico Matteoni and
                  Andrea Cossu and
                  Claudio Gallicchio and
                  Vincenzo Lomonaco and
                  Davide Bacciu},
  title        = {Continual Learning for Human State Monitoring},
  booktitle    = {30th European Symposium on Artificial Neural Networks, Computational
                  Intelligence and Machine Learning, {ESANN}},
  year         = {2022}
}

@article{yu2022coca,
  title={CoCa: Contrastive Captioners are Image-Text Foundation Models},
  author={Jiahui Yu and Zirui Wang and Vijay Vasudevan and Legg Yeung and Mojtaba Seyedhosseini and Yonghui Wu},
  journal={Trans. Mach. Learn. Res.},
  year={2022}
}

@inproceedings{zeng2022transformers,
  title={Are Transformers Effective for Time Series Forecasting?},
  author={Zeng, Ailing and Chen, Muxi and Zhang, Lei and Xu, Qiang},
  booktitle = {Proceedings of the 37 AAAI Conference on Artificial Intelligence},
articleno = {1248},
numpages = {8},
  year={2023}
}

@inproceedings{ramasesh2022effect,
  title={Effect of Model and Pretraining Scale on Catastrophic Forgetting},
  author={Ramasesh, Vinay and et al.},
  booktitle={International Conference on Learning Representations (ICLR)},
  year={2022}
}

@article{runge2023causal,
  title={Causal inference for time series},
  author={Runge, Jakob and Gerhardus, Andreas and Varando, Gherardo and Eyring, Veronika and Camps-Valls, Gustau},
  journal={Nat. Rev. Earth \& Environment},
  volume={4},
  number={7},
  pages={487--505},
  year={2023}
}

@inproceedings{zhai2023investigating,
  title={Investigating the Catastrophic Forgetting in Multimodal Large Language Model Fine-Tuning},
  author={Yuexiang Zhai and Shengbang Tong and Xiao Li and Mu Cai and Qing Qu and Yong Jae Lee and Yi Ma},
  booktitle={First Conference on Parsimony and Learning (CPAL)},
  year={2024}
}

@inproceedings{gruver2023large,
author = {Gruver, Nate and Finzi, Marc and Qiu, Shikai and Wilson, Andrew Gordon},
title = {Large language models are zero-shot time series forecasters},
year = {2023},
booktitle = {Proceedings of the 37th International Conference on Neural Information Processing Systems},
articleno = {861},
numpages = {14},
}

@article{ao2023continual,
    author = {Ao, SI and Fayek, H},
    title = {Continual Deep Learning for Time Series Modeling},
    journal = {Sensors},
    volume={23},
    number={16},
    year = {2023}
}

@inproceedings{Gunasekara2023,
  title     = {Survey on Online Streaming Continual Learning},
  author    = {Gunasekara, Nuwan and Pfahringer, Bernhard and Gomes, Heitor Murilo and Bifet, Albert},
  booktitle = {Proceedings of the Thirty-Second International Joint Conference on
               Artificial Intelligence, {IJCAI-23}},
  pages     = {6628--6637},
  year      = {2023},
  note      = {Survey Track}
}

@inproceedings{ilbert2024samformer,
author = {Ilbert, Romain and Odonnat, Ambroise and Feofanov, Vasilii and Virmaux, Aladin and Paolo, Giuseppe and Palpanas, Themis and Redko, Ievgen},
title = {SAMformer: unlocking the potential of transformers in time series forecasting with sharpness-aware minimization and channel-wise attention},
year = {2024},
booktitle = {Proceedings of the 41st International Conference on Machine Learning},
articleno = {841},
numpages = {31},
series = {ICML'24}
}

@article{ansari2024chronos,
title={Chronos: Learning the Language of Time Series},
author={Ansari, A. F. and et. al.},
journal={Transactions on Machine Learning Research},
year={2024}
}

@inproceedings{woo2024unified,
author = {Woo, Gerald and Liu, Chenghao and Kumar, Akshat and Xiong, Caiming and Savarese, Silvio and Sahoo, Doyen},
title = {Unified training of universal time series forecasting transformers},
year = {2024},
booktitle = {Proceedings of the 41st International Conference on Machine Learning},
articleno = {2178},
numpages = {25}
}

@ARTICLE{wang2024comprehensive,
author={Wang, Liyuan and Zhang, Xingxing and Su, Hang and Zhu, Jun},
journal={ IEEE Transactions on Pattern Analysis \& Machine Intelligence },
title={{A Comprehensive Survey of Continual Learning: Theory, Method and Application }},
year={2024},
volume={46},
number={08},
ISSN={1939-3539},
pages={5362-5383},
publisher={IEEE Computer Society}
}

@article{schneider2024foundation,
  title={Foundation Models: A New Paradigm for Artificial Intelligence},
  author={Schneider, J. and Meske, C. and Kuss P.},
  journal={Business \& Information Systems Engineering},
volume={66},
page={221–231},
  year={2024}
}

@inproceedings{Liang24,
author = {Liang, Yuxuan and Wen, Haomin and Nie, Yuqi and Jiang, Yushan and Jin, Ming and Song, Dongjin and Pan, Shirui and Wen, Qingsong},
title = {Foundation Models for Time Series Analysis: A Tutorial and Survey},
year = {2024},
isbn = {9798400704901},
publisher = {ACM},
booktitle = {Proceedings of the 30th ACM SIGKDD Conference on Knowledge Discovery and Data Mining},
pages = {6555–6565},
numpages = {11},
series = {KDD '24}
}

@inproceedings{das2024timesfm,
author = {Das, Abhimanyu and Kong, Weihao and Sen, Rajat and Zhou, Yichen},
title = {A decoder-only foundation model for time-series forecasting},
year = {2024},
booktitle = {Proceedings of the 41st International Conference on Machine Learning},
articleno = {404},
numpages = {20},
location = {Vienna, Austria},
series = {ICML'24}
}

@inproceedings{besnard2024continual,
  title={Continual Learning for Time Series Forecasting: A First Survey},
  author={Besnard, Quentin and Ragot, Nicolas},
  booktitle={The 10th International Conference on Time Series and Forecasting},
  year={2024},
  series = {MDPI Engineering Proceedings}
}

@Article{Bayle2024,
AUTHOR = {Bayle, Raphaël and Reyboz, Marina and Lomet, Aurore and Cook, Victor and Mermillod, Martial},
TITLE = {Continuously Learning Prediction Models for Smart Domestic Hot Water Management},
JOURNAL = {Energies},
VOLUME = {17},
YEAR = {2024},
NUMBER = {18},
ARTICLE-NUMBER = {4734}
}

@inproceedings{aksu2024gift,
title={{GIFT}-Eval: A Benchmark for General Time Series Forecasting Model Evaluation},
author={Taha Aksu and Gerald Woo and Juncheng Liu and Xu Liu and Chenghao Liu and Silvio Savarese and Caiming Xiong and Doyen Sahoo},
booktitle={NeurIPS Workshop on Time Series in the Age of Large Models},
year={2024}
}

@INPROCEEDINGS{Liu25,
  author={Liu, Jia and Jinguo, Cheng and Fang, Xia and Ma, Zhenyuan and Wu, Yuankai},
  booktitle={2025 International Joint Conference on Neural Networks (IJCNN)}, 
  title={Evaluating Temporal Plasticity in Foundation Time Series Models for Incremental Fine-Tuning}, 
  year={2025}
  }

@article{luo2025empirical,
 author={Luo, Yun and Yang, Zhen and Meng, Fandong and Li, Yafu and Zhou, Jie and Zhang, Yue},
  journal={IEEE Transactions on Audio, Speech and Language Processing}, 
  title={An Empirical Study of Catastrophic Forgetting in Large Language Models During Continual Fine-Tuning}, 
  year={2025},
  volume={33},
  pages={3776-3786},
}

@article{haque2025catastrophic,
  title={Catastrophic Forgetting in LLMs: Comparative Analysis},
  author={Haque, Naimul},
  journal={arXiv:2504.01241},
  year={2025}
}

@inproceedings{
auerTiRexZeroShotForecasting2025,
title={TiRex: Zero-Shot Forecasting Across Long and Short Horizons with Enhanced In-Context Learning},
author={Andreas Auer and Patrick Podest and Daniel Klotz and Sebastian B{\"o}ck and G{\"u}nter Klambauer and Sepp Hochreiter},
booktitle={The 39th Annual Conference on Neural Information Processing Systems},
year={2025}
}

@article{graf2025flowstate,
title={FlowState: Sampling Rate Invariant Time Series Forecasting},
  author={Lars Graf and Thomas Ortner and Stanisław Woźniak and Angeliki Pantazi},
  journal={arXiv preprint arXiv:2508.05287},
  year={2025}
}

@inproceedings{karaouli2025foundational,
  title={How Foundational are Foundation Models for Time Series Forecasting?},
  author={Karaouli, Nouha and Coquenet, Denis and Fromont, Elisa and Mermillod, Martial and Reyboz, Marina},
  booktitle={Recent Advances in Time Series Foundation Models Have We Reached the 'BERT Moment'?},
  year={2025}
}

@article{ansari2025chronos2,
  title        = {Chronos-2: From Univariate to Universal Forecasting},
  author       = {Abdul Fatir Ansari and et. al.},
  year         = {2025},
   journal      = {arXiv preprint arXiv:2510.15821},
  year         = {2025}
}

@article{Shi26,
author = {Shi, Haizhou and Xu, Zihao and Wang, Hengyi and Qin, Weiyi and Wang, Wenyuan and Wang, Yibin and Wang, Zifeng and Ebrahimi, Sayna and Wang, Hao},
title = {Continual Learning of Large Language Models: A Comprehensive Survey},
year = {2026},
publisher = {ACM},
volume = {58},
number = {5},
journal = {ACM Comput. Surv.},
month = nov,
articleno = {120},
numpages = {42}
}
\newpage

\end{document}